\def\paperID{4626}
\def\confName{ICCV}
\def\confYear{2025}

\newif\ifreview 
\newif\ifarxiv \newcommand{\arxiv}{\arxivtrue}
\newif\ifcamera 
\newif\ifrebuttal 

\arxiv
\pdfoutput=1
\documentclass[10pt,twocolumn,letterpaper]{article}
\ifreview \usepackage[review]{cvpr} \fi
\ifarxiv \usepackage[pagenumbers]{cvpr} \fi
\ifrebuttal \usepackage[rebuttal]{cvpr} \fi
\ifcamera \usepackage{cvpr} \fi

\usepackage{graphicx}	
\usepackage{amsmath}	
\usepackage{amssymb}	
\usepackage{booktabs}
\usepackage{times}
\usepackage{microtype}
\usepackage{epsfig}
\usepackage{caption}
\usepackage{float}
\usepackage{placeins}
\usepackage{color, colortbl}
\usepackage{stfloats}
\usepackage{enumitem}
\usepackage{tabularx}
\usepackage{xstring}
\usepackage{multirow}
\usepackage{xspace}
\usepackage{url}
\usepackage{subcaption}
\usepackage{xcolor}
\usepackage{wrapfig}

\usepackage[hang,flushmargin]{footmisc}

\ifcamera \usepackage[accsupp]{axessibility} \fi

% ...

\ifarxiv  \fi

\newcommand{\R}[1]{{%
    \textbf{%
        \ifstrequal{#1}{1}{\textcolor{red}{R#1}}{%
        \ifstrequal{#1}{2}{\textcolor{blue}{R#1}}{%
        \ifstrequal{#1}{3}{\textcolor{magenta}{R#1}}{%
        \ifstrequal{#1}{4}{\textcolor{teal}{R#1}}{%
                           \textcolor{cyan}{R#1}%
        }}}}%
    }%
}}

\usepackage{xr-hyper}

\makeatletter
\newcommand*{\addFileDependency}[1]{
  \typeout{(#1)}
  \@addtofilelist{#1}
  \IfFileExists{#1}{}{\typeout{No file #1.}}
}

\makeatother

\definecolor{cvprblue}{rgb}{0.21,0.49,0.74}
\usepackage[pagebackref,breaklinks,colorlinks,allcolors=cvprblue]{hyperref}
\usepackage[capitalize]{cleveref}
\crefname{section}{Sec.}{Secs.}
\crefname{table}{Table}{Tables}
\crefname{figure}{Fig.}{Figs.}

\ifarxiv \crefname{appendix}{App.}{Apps.}
\else \crefname{appendix}{Suppl.}{Suppls.} \fi

\author{Mengwei Xie\textsuperscript{\rm 1}, \hspace{0.75em}
Shuang Zeng\textsuperscript{\rm 1, 2},\hspace{0.75em} Xinyuan Chang\textsuperscript{\rm 1}, \hspace{0.75em}Xinran Liu\textsuperscript{\rm 1}, \\ Zheng Pan\textsuperscript{\rm 1}, \hspace{0.75em}Mu Xu\textsuperscript{\rm 1}, \hspace{0.75em}Xing Wei\textsuperscript{\rm 2\dag}\\
\textsuperscript{\rm 1}Amap, Alibaba Group \hspace{0.5em}\textsuperscript{\rm 2}Xi’an Jiaotong University    \hspace{0.5em} \\
\tt\small \{xiemengwei.xmw, changxinyuan.cxy,  tom.lxr, panzheng.pan, xumu.xm\}@alibaba-inc.com, \\
\tt\small zengshuang@stu.xjtu.edu.cn, weixing@mail.xjtu.edu.cn
}

\begin{document}
\title{SeqGrowGraph: Learning Lane Topology as a Chain of Graph Expansions}

\maketitle
\renewcommand\thefootnote{}\footnotetext{$^\dag$ Corresponding author.}

\begin{abstract}


Accurate lane topology is essential for autonomous driving, yet traditional methods struggle to model the complex, non-linear structures—such as loops and bidirectional lanes—prevalent in real-world road structure. We present \textbf{SeqGrowGraph}, a novel framework that learns lane topology as a chain of graph expansions, inspired by human map-drawing processes. Representing the lane graph as a directed graph $G=(V,E)$, with intersections ($V$) and centerlines ($E$), SeqGrowGraph incrementally constructs this graph by introducing one vertex at a time. At each step, an adjacency matrix ($A$) expands from $n \times n$ to $(n+1) \times (n+1)$ to encode connectivity, while a geometric matrix ($M$) captures centerline shapes as quadratic Bézier curves. The graph is serialized into sequences, enabling a transformer model to autoregressively predict the chain of expansions, guided by a depth-first search ordering. Evaluated on nuScenes and Argoverse 2 datasets, SeqGrowGraph achieves state-of-the-art performance.

\end{abstract}
\section{Introduction}
\label{sec:intro}

The rise of autonomous driving has heightened the need for precise road structure modeling to ensure safe and efficient navigation~\cite{DBLP:conf/cvpr/PittnerJC24, DBLP:conf/aaai/ZhengZMGLHPY24, DBLP:conf/eccv/LiHYZS24}. Road structures encompass explicit features such as centerlines, road edges, and traffic signs, as well as implicit topological relationships, notably the connectivity between centerlines. Constructing lane graphs is crucial for representing these complex structures, facilitating vehicle decision-making and path planning in dynamic environments.

Traditional lane graph generation methods fall into {detection-based} and {generation-based} approaches, each with notable limitations. Detection-based methods can be categorized into three types: 1) Pixel-level approaches, such as HDMapNet~\cite{HDMapNet}, segment lane structures in bird's-eye view (BEV) space on a per-pixel basis but struggle to infer global topology. 2) Piece-level methods~\cite{TopoNet,TopoNetv2,STSU} detect centerlines in fragmented segments based on topological points, often resulting in discontinuities and misalignments (Figure~\ref{problem}). 3) Path-level methods, such as LaneGAP~\cite{LaneGAP}, aim to mitigate discontinuities by detecting lane paths but introduce redundancies and alignment errors due to heavy reliance on post-processing.

\begin{figure}[tp]
    \centering
    \includegraphics[width=0.9\linewidth]{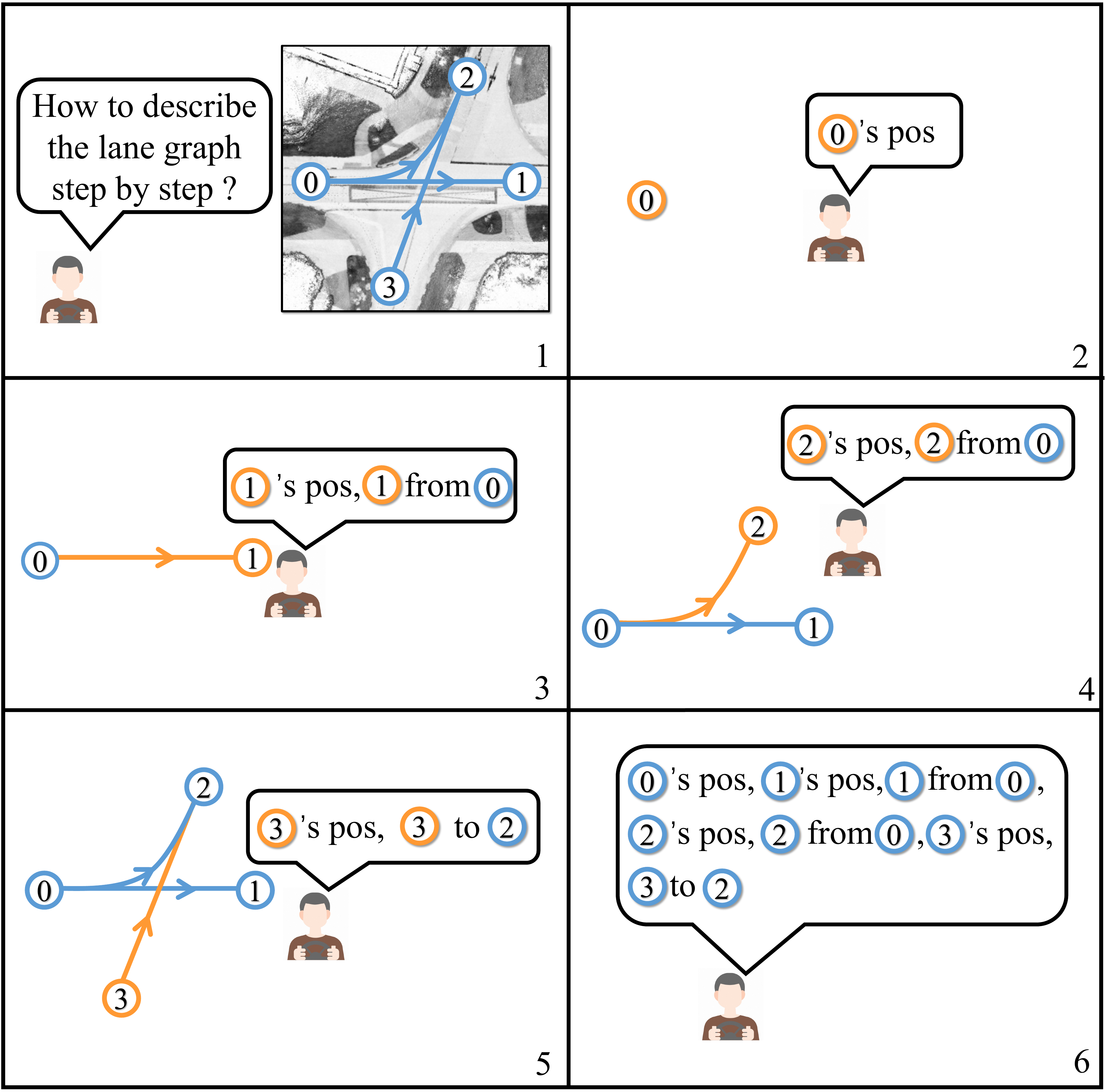}
    \caption{\textbf{Describing the lane graph step by step.} Starting from an initial point, it is necessary to determine the location of the next point and establish whether there is a ``from'' or ``to'' relationship with the existing points. By repetitively carrying out this process, a complete description of the lane graph can be obtained. The abbreviation ``pos'' in the diagram stands for ``position''.}
    \label{people}
\end{figure}

Within the widely applied paradigm of sequence generation~\cite{Liang_2025_CVPR,wei2023autoregressive,Bai_2024_CVPR,chen2021pix2seq}, generative models are regarded as offering a more direct method for inferring lane graphs by capturing hidden structural patterns. 

However, formulating lane graphs as sequences poses challenges and existing generative models have certain limitations. For instance, RNTR~\cite{roadnet-iccv} converts lane graph into a Directed Acyclic Graph (DAG), which brings sequence redundancy in pre-processing and post-processing and fails to express loops and bidirectional lanes. LaneGraph2Seq~\cite{roadnet-aaai}, another generation-based method, generates the junction positions and shapes of centerlines separately, missing a holistic understanding of the network topology and struggling with integrated representation. And DAG-based methods fail to express loops and bidirectional lanes. Existing methodologies for graph-to-sequence conversion tend to excessively deconstruct the intrinsic graph structure, leading to transformations that disrupt the original topology.


\textit{Wait—how do humans construct lane graphs?} Instead of processing an entire graph at once, people naturally start from a single node and progressively add new elements, continuously expanding the graph by establishing connections with previously defined nodes (Figure~\ref{people}). This intuitive approach suggests a novel formulation:
1) Modeling lane graph construction as a chain of stepwise expansion process. 2) Using sequence modeling to capture incremental adjacency matrix updates.

Inspired by this observation, we propose \textbf{SeqGrowGraph}, a novel framework for incrementally constructing a lane \textbf{Graph} via \textbf{Seq}uence \textbf{Grow}th.
SeqGrowGraph represents a lane graph as a directed graph $G = (V, E)$, where $V$ denotes intersections or key topological nodes, and $E$ represents centerlines, defining connectivity between nodes. SeqGrowGraph models lane topology as a chain of graph expansions:
\begin{itemize}
    \item A new node is introduced with its spatial position.
    \item The adjacency matrix $A$ expands from $n \times n$ to $(n+1) \times (n+1)$ to encode connectivity, with the upper triangle representing incoming edges (``from'') and the lower triangle representing outgoing edges (``to'').
    \item A geometric matrix $M$ updates the centerline shape, represented as quadratic Bézier curves.
    \item A transformer-based generative model autoregressively predicts these expansions using a depth-first search ordering.
\end{itemize}


Unlike DAG-based approaches, SeqGrowGraph flexibly models complex, real-world road structures, including loops, bidirectional lanes, and non-trivial topologies, without excessive pre-processing or post-processing. Our key contributions are as follows:
\begin{itemize}
    \item We introduce SeqGrowGraph, a novel framework that reformulates lane graph generation as a sequential expansion process, leveraging adjacency matrix updates and geometric modeling.
    \item We develop an autoregressive transformer-based model that progressively constructs lane graphs, overcoming the limitations of DAG-based methods.
    \item We achieve state-of-the-art performance on large-scale datasets (nuScenes, Argoverse 2), demonstrating superior topological accuracy and network completeness.
\end{itemize}

SeqGrowGraph offers a practical and effective solution for lane topology modeling, aligning closely with the natural human approach to map construction.

\begin{figure}[tp]
    \centering
    \includegraphics[width=0.9\linewidth]{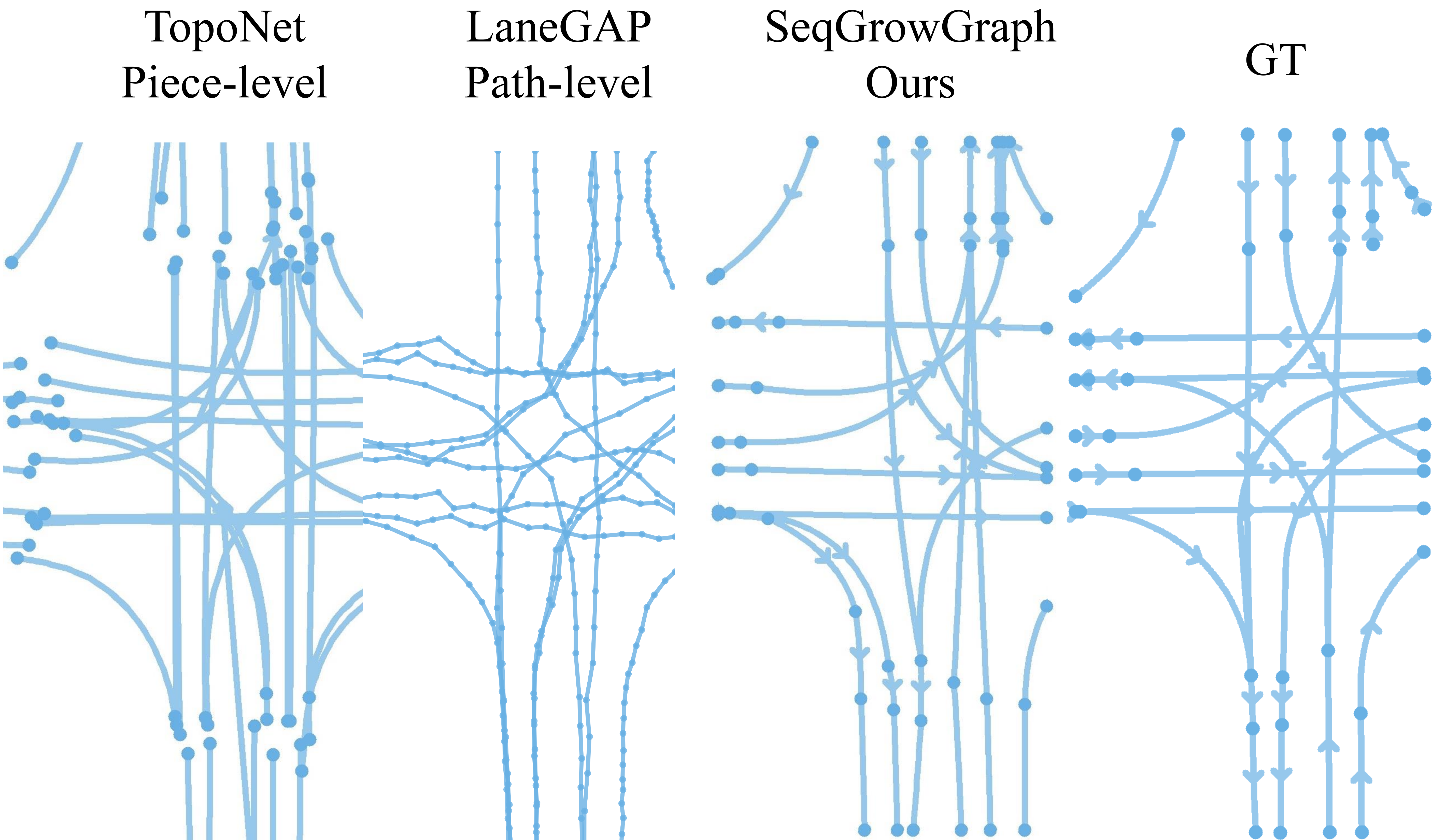}
    \caption{\textbf{Visualization comparison of lane graph results from different methods.} This figure illustrates the inference results using the official models of TopoNet~\cite{TopoNet} and LaneGAP~\cite{LaneGAP}. It can be observed that TopoNet, by detecting centerlines separately, results in discontinuities between the centerlines. In contrast, LaneGAP, which detects at the path level, tends to confuse multiple lanes and duplicate paths. Both methods rely heavily on extensive post-processing, failing to generate continuous centerline results in a single step.}
    \label{problem}
\end{figure}

\section{Related work}

\begin{figure*}[tp]
    \centering
    \includegraphics[width=\linewidth]{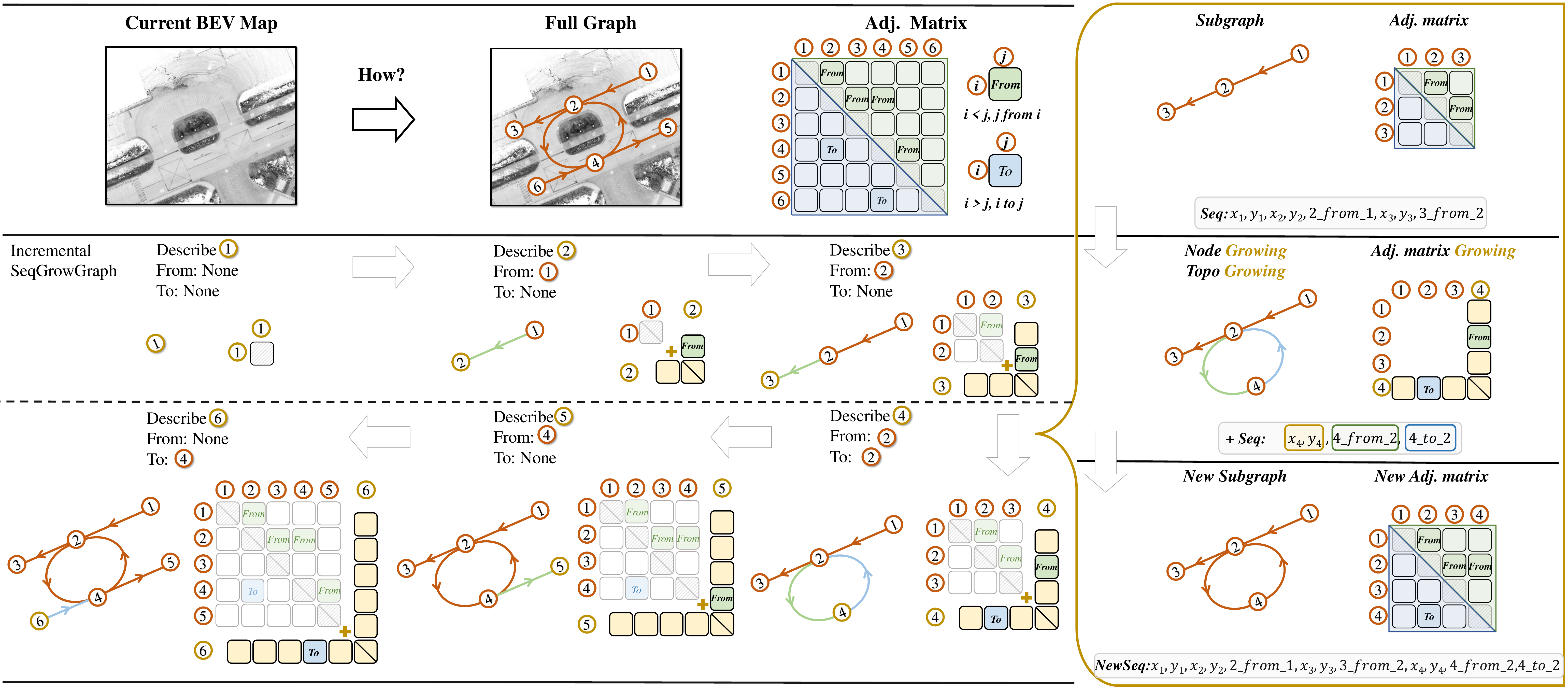}
    \caption{\textbf{Expansion process of the adjacency matrix.} Each node corresponding to an intersections or key topological node and edges representing centerlines. At each step, we describe a new node along with the its connected centerlines. We focus only on the topological relationships between the newly introduced node and the existing nodes. To facilitate better formatting into a sequence, we use the new node as the subject to describe its topology. The terms ``from" and ``to" respectively indicate which nodes the new node can come ``from" and go ``to".
    For example, when describing a new node \footnotesize{\textcircled{\scriptsize{$n$}}}\normalsize\, we specify its position and connections to existing nodes, corresponding to the non-zero elements in the \( n \)-th row and \( n \)-th column of the adjacency matrix. By expanding the adjacency matrix row by row and column by column, we generate subgraphs that are incrementally combined to form the complete graph.
    }
    \label{fig1}
\end{figure*}

\subsection{Online HD map construction} 
Online high-definition map construction enables accurate road geometry modeling for autonomous navigation, with contemporary approaches~\cite{vectormapnet,HDMapNet,MapTR, MapTRv2,HIMap,zeng2024driving} focusing on BEV-space localization of 2D map elements. Initial approaches conceptualized this task as a semantic segmentation problem, as demonstrated in foundational works like HDMapNet~\cite{HDMapNet}, which necessitated computationally intensive post-processing procedures to convert pixel-level segmentation outputs into usable vector formats. To address this limitation, VectorMapNet~\cite{vectormapnet} has focused on developing end-to-end frameworks for vectorized map learning. This approach employs a two-phase hierarchical architecture utilizing topological keypoint detection, implementing an initial coarse prediction stage followed by geometric refinement processes to achieve precise vectorized outputs. Notably, MapTR~\cite{MapTR} achieves competitive performance using only PV image inputs, eliminating LiDAR dependency. 3D lane detection in autonomous driving follows two primary paradigms: BEV-based and BEV-free approaches.
BEV-based methods~\cite{DBLP:conf/cvpr/PittnerJC24,DBLP:journals/ral/LiHGYYWZZ24,DBLP:conf/iccv/YaoY0J23} transform perspective view (PV) images into bird's-eye view (BEV) representations using camera intrinsics and ground coordinates to extract height-aware BEV features for accurate 3D lane modeling. BEV-free methods~\cite{DBLP:conf/aaai/ZhengZMGLHPY24,DBLP:conf/iccv/LuoZYKZCL23,DBLP:conf/icra/BaiCFPLC23} eliminates view transformation by operating directly in 3D space, leveraging perspective view information for direct 3D lane detection and reconstruction.

\subsection{Language models for graph}

Language models (LMs) have exhibited impressive capabilities not only in processing pure text but also across diverse application scenarios for autonomous driving~\cite{zeng2025futuresightdrivethinkingvisuallyspatiotemporal,chang2025drivingrulesbenchmarkintegrating}. A significant challenge in graph fields is leveraging LMs to effectively understand and reason with graph information. How to encode graphs into learnable sequences according to specific rules is essential. Researchers~\cite{instructGLM,DBLP:conf/iclr/MengZLS0WL22,DBLP:journals/tmlr/Huang0M024} have demonstrated that language models possess substantial potential for learning from graphs. Lane graph, which are inherently graph-like with their interdependent nodes and edges, can significantly benefit from the contextual understanding and pattern recognition capabilities of language models. The integration of these models with graph structures shows great promise for advancing the comprehension of lane graph.

\subsection{Lane connectivity topology}

 Lane topology understanding is a crucial aspect of comprehensive road comprehension. STSU~\cite{STSU} is the first to introduce this task, predicting lane centerlines and their connectivity. TopoNet~\cite{TopoNet} views the lane  topology as a graph, employing an end-to-end framework with a graph neural network architecture to learn the complete road structure. These methods treat the task of extracting centerlines and topology as two separate components, which can lead to a lack of  coherence and misalignment between centerlines. LaneGAP~\cite{LaneGAP} tackles these challenges by directly recognizing paths as units, accounting for existing overlapping centerlines between paths. However, this approach still requires stringent processing to remove duplicates.
RNTR~\cite{roadnet-iccv} and LaneGraph2Seq~\cite{roadnet-aaai} view the lane graph as a DAG and establish a bidirectional transformation between the DAG and sequences, leveraging language models to generate sequences from PV images. RNTR~\cite{roadnet-iccv}  requires converting the DAG into a directed forest first, which involves complex pre-processing and post-processing steps, and the sequence information is redundant. On the other hand, LaneGraph2Seq~\cite{roadnet-aaai} separates the learning of nodes and edges in the DAG by learning nodes first, followed by learning edges, which does not align with the language model's step-by-step reasoning approach. Addressing these challenges through a holistic approach could enhance the coherence and efficiency of lane graph modeling.

\label{sec:related}

\section{Method}

In this section, we will introduce the mathematical modeling of the lane graph in Section~\ref{3.1_math}, the method of serializing the lane graph in Section~\ref{3.2_lane_graph}, and introduce the model architecture and objective in Section~\ref{3.3_model}.

\subsection{Mathematical modeling of lane graph}
\label{3.1_math}


The lane graph formed by centerlines can be modeled as a directed graph \( G = (V, E) \), where \( V \) denotes a set of vertices and \( E \) represents a set of edges. The vertices \( V = \{ v_1, v_2, v_3, \ldots \} \) correspond to intersections or key topological nodes \( \{ 1, 2, 3, \ldots \} \), where each vertex \( v_n = (x_n, y_n) \) specifies its geographic position. The edges \( E \) signify the lane lines connecting these intersections. To represent \( E \), we employ an adjacency matrix \( A \) alongside an additional matrix \( M \).
The adjacency matrix \( A \) indicates the connectivity between the vertices and
the matrix \( M \) stores the shape of the centerlines that connect the intersections represented by the vertices.

\begin{itemize}

 \item \(A\) :  The adjacency matrix \( A \) indicates the connectivity between the vertices. An entry \( A[i][j] \) typically denotes whether there is a directed edge (directed centerline) from vertex \( v_i \) to vertex \( v_j \) without passing through other edges. Specifically,
    \[
    A(i, j) = 
    \begin{cases} 
    1, & \text{if there is a  edge  between  } v_i \text{ and } v_j \\ 
    0, & \text{otherwise}
    \end{cases}
    \]
    Since lane graph is directed, \( A(i, j) \neq A(j, i)  \)

 \item \(M\):  The matrix \( M \) stores the shape of the centerlines that connect the intersections represented by the vertices. Specifically,  \( M(i, j) \) represents the centerline position from intersection \( i \) to intersection \( j \), expressed using a quadratic Bézier function. This function includes the starting point \( v_i \) and the endpoint \( v_j \), whose positions are already stored in \( V \), along with a middle control point at position \((\sigma_x^{ij}, \sigma_y^{ij})\). Therefore, we define
  \[
    M(i, j) = 
    \begin{cases} 
    (\sigma_x^{ij}, \sigma_y^{ij}), & \text{if  }A(i, j)=1 \\ 
    \emptyset, & \text{if  }A(i, j)=0
    \end{cases}
    \]
 
Together, \( G = (V, A,M) \) provide a comprehensive representation of the lane graph, encapsulating both the connectivity between intersections and the geometric details of the centerlines.

\end{itemize}

\subsection{Serializing the lane graph}
\label{3.2_lane_graph}




In this section, we introduce the method of modeling lane graph as sequences. 
Previous research has primarily considered the lane graph as a Directed Acyclic Graph (DAG), taking into account that most roads are one-way  and acyclic. However, real-world road infrastructure is  complex, even in the local map surrounding a vehicle, the road graph may contain loops and bidirectional single-lane roads, which are issues not considered by previous work. We propose a novel modeling approach called SeqGrowGraph. 

As previously defined, \( v_n = (x_n, y_n) \) denotes the position of intersection node \( n \), and \( M \) stores the positions of the centerlines, with \( M(i, j) \) representing the control point of the Bézier curve for the centerline between nodes \( i \) and \( j \). 

We define \( S_n \) as the sequence that represents the subgraph formed by the first \( n \) nodes. Let \( F_n \) denote the set of all centerlines from which node \( n \) can originate, and \( T_n \) the set of all centerlines to which node \( n \) can proceed. Here, the \( + \) symbol indicates sequence concatenation. Our approach grows as follows: \( S_n \)  is the concatenation of the preceding sequence and the new subsequences from node \( n \). The addition operation denotes sequence concatenation.

\begin{gather*}
S_{n} = S_{n-1} + (v_n + F_n + T_n) \\
\text{where} \quad v_n = (x_n, y_n)
\end{gather*}

The sets \( F_n \) and \( T_n \), which represent the centerlines connecting the \( n \)-th node to the preceding \( n-1 \) nodes, are expressed as:

\begin{gather*}
F_n = \sum_{k=0}^{n-1} M(k, n) \\
T_n = \sum_{k=0}^{n-1} M(n, k)
\end{gather*}

Assuming there are a total of $N$ topological points, the final 
$S_N$  represents the ultimate sequence, which forms a complete description of the lane graph.

Figure~\ref{fig1} illustrates our step-by-step process for describing a graph as a sequence. This progressive expansion of the adjacency matrix and subgraph results in a comprehensive description of the lane graph. The left side of Figure~\ref{fig1} presents the incremental depiction of subgraphs, comprising six steps and corresponding subgraphs.
In each step, we describe one node. As for the order in which the nodes should be described at each step, our method utilizes depth-first search.

The right side of Figure~\ref{fig1} uses the introduction of \footnotesize{\textcircled{\scriptsize{$4$}}}\normalsize\ as an example to demonstrate this incremental subgraph description. 

1. Before introducing \footnotesize{\textcircled{\scriptsize{$4$}}}\normalsize\, the adjacency matrix \( A \) includes only the first \( 3 \) rows and \( 3 \) columns, capturing the existing topological information. 

2. After specifying the position \(x_4, y_4\) of node \footnotesize{\textcircled{\scriptsize{$4$}}}\normalsize\, we describe its topology by considering the first 4 rows and first 4 columns of matrix A.
The non-zero elements in the \( 4 \)-th row indicate the nodes to which \footnotesize{\textcircled{\scriptsize{$4$}}}\normalsize\  can directly go   ``to"; for instance, \( A(4,2) = 1 \) signifies that \footnotesize{\textcircled{\scriptsize{$4$}}}\normalsize\ can go to \footnotesize{\textcircled{\scriptsize{$2$}}}\normalsize\ without passing any other node. 
We use the notation \(4\_to\_2\) to represent the corresponding centerline. 
Conversely, the non-zero elements in the \( 4 \)-th column indicate the nodes that \footnotesize{\textcircled{\scriptsize{$4$}}}\normalsize\ can come ``from"; for instance, \( A(2,4) = 1 \) means that \footnotesize{\textcircled{\scriptsize{$4$}}}\normalsize\ can come directly from \footnotesize{\textcircled{\scriptsize{$2$}}}\normalsize\ . We use \(4\_from\_2\) to denote the corresponding centerline.

3. The new sequence, formed by concatenating the previous sequence with the information of the current node, represents the subgraph of all nodes present so far. This process is repeated iteratively, ultimately yielding a final sequence that represents the complete lane graph.

\begin{figure}[tp]
    \centering
    \includegraphics[width=0.9\linewidth]{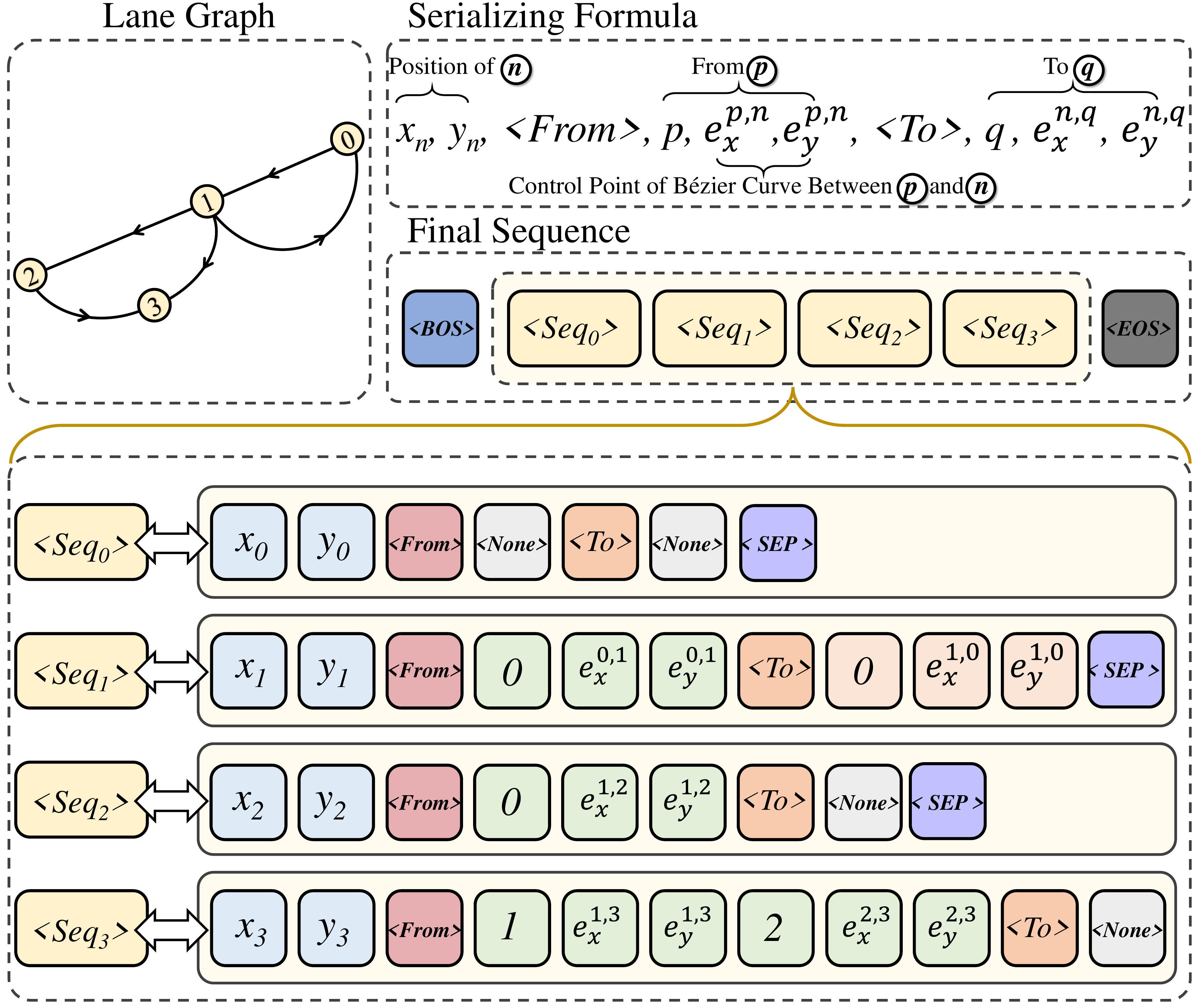}
    
    \caption{\textbf{The demonstration of graph serialization by SeqGrowGraph.} Each node corresponds to a sequence, and the sequences are connected by $<SEP>$, which together represent   final lane graph.}
    \label{example}
\end{figure}


Figure~\ref{example} illustrates the sequence structure of SeqGrowGraph. Each node's sequence includes its shape, connected node indices, and the Bézier control points of the connecting centerlines. By concatenating these sequences, we generate the target sequence. This serialized representation enables a more flexible expression of the topology, while maintaining clarity and conciseness and preserving essential graph-level information.

Once the complete sequential representation for the lane graph has been generated, the next step is to express the positions and Bézier curve control points in a discretized form, excluding the special tokens and indices. This involves dividing the coordinates within the ego vehicle's system into discrete bins, allowing us to assign a corresponding integer number to each position. It is important to note that the range of special tokens is greater than the total number of bins.

\begin{figure}[tp]
    \centering
    \includegraphics[width=\linewidth]{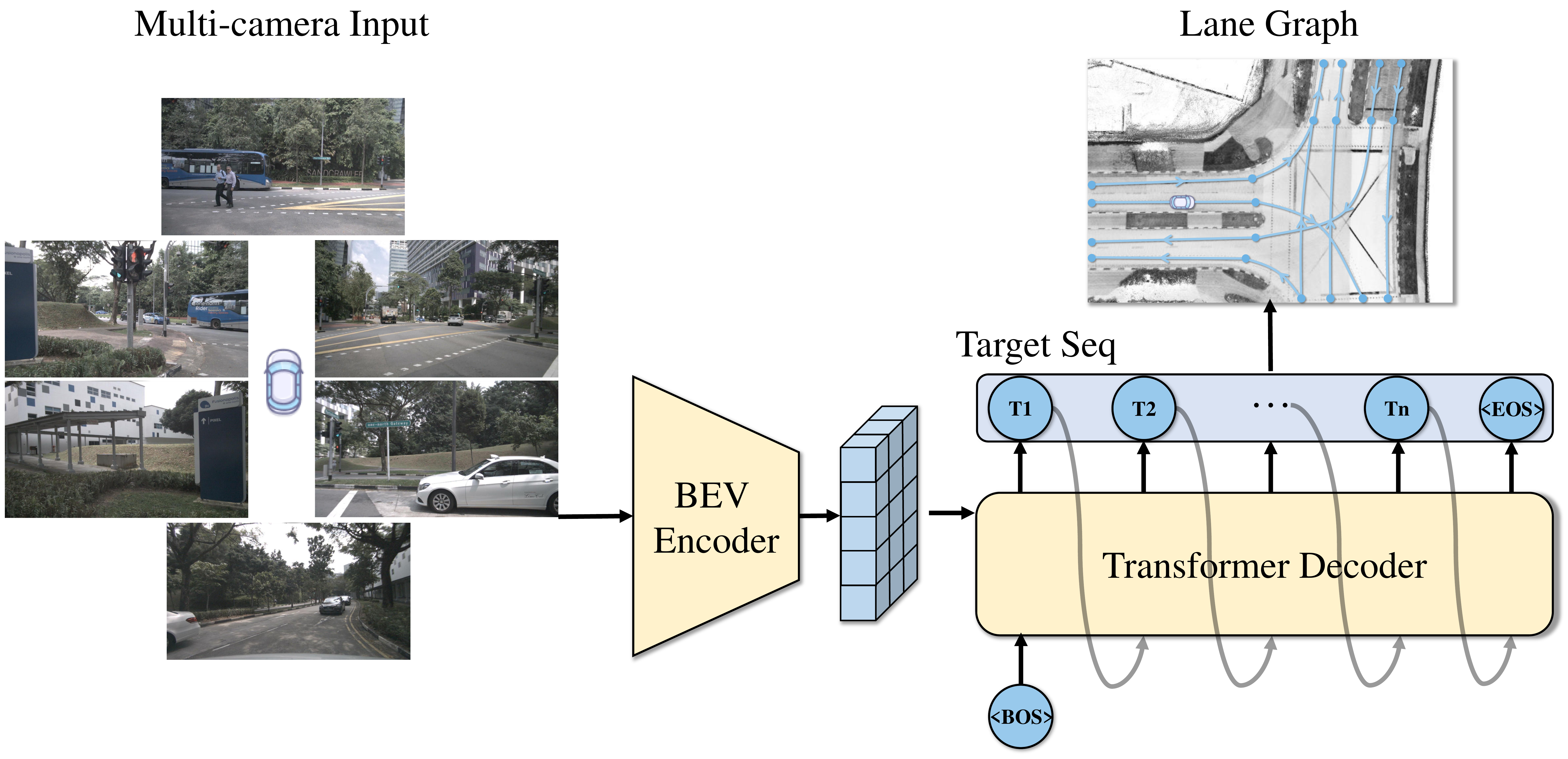}
    
    \caption{\textbf{Model architecture and inference process.} We employ a BEV-encoder to transform surrounding camera perspective view (PV) images into BEV features. Subsequently, a Transformer decoder generates tokens of the target sequence based on the BEV features and the beginning of sequence $<BOS>$. All the tokens generated  represent a complete lane graph.}
    \label{fig2}
\end{figure}

\subsection{Model architecture and objective}
\label{3.3_model}

\paragraph{Architecture} Following prior studies ~\cite{roadnet-iccv,roadnet-aaai}, our method commences with the use of  BEV encoder LSS~\cite{lss} to project features extracted from surrounding images onto the BEV plane. Following the conversion of the lane graph into a sequence, a transformer decoder is employed to generate the sequence token by token. The comprehensive architecture of our approach is depicted in Figure~\ref{fig2}.

\paragraph{Obejctive} We define the final target sequence \( S = (x_1, x_2, \ldots, x_T) \) for each sample and have \(\hat{x}\)  represent the predicted value.
 During the training phase, the loss function is expressed using maximum likelihood estimation as follows:

\[
\mathcal{L} = -\left( \frac{1}{T} \sum_{t=1}^{T} \log p(\hat{x}_t \mid x_1 , x_2 , \ldots, x_{t-1} ) \right)
\]
 This loss function measures the model's performance by taking the negative logarithm of the probability of predicting each subsequent step given the previous true token.

\label{sec:method}

\section{Experiments}




\subsection{Dataset}
We evaluate our method using two widely-used autonomous driving research datasets: nuScenes~\cite{nuScenes} and Argoverse 2~\cite{Argoverse2}. The nuScenes dataset comprises 1,000 scenes  with six cameras offering a 360° field of view. Due to overlapping lanes in the default nuScenes split, PON~\cite{pon} redistributes the train/validation sequences to eliminate overlap, allowing us to test our approach on both the default and PON splits. Consistent with past studies~\cite{roadnet-iccv, roadnet-aaai}, we input all surrounding camera images and  Bird's Eye View (BEV) target ranges from -48m to 48m on the x-axis and -30m to 30m on the y-axis. The Argoverse 2 dataset contains 1,000 sequences and includes seven cameras, also providing a 360° field of view. For this dataset, we use camera images as input, with the BEV target range from -30m to 30m on the x-axis and -15m to 15m on the y-axis.

\subsection{Implementation}
Following~\cite{roadnet-aaai} We employ an LSS encoder~\cite{lss} based on ResNet50 as the BEV encoder, initially training this encoder on a centerline segmentation task. For the sequence decoder, we use a 6-layer transformer. The training process is conducted on 4 NVIDIA A100 GPUs with  batch size of 4x18. Specifically, the nuScenes dataset is trained for 400 epochs and the Argoverse 2 dataset is trained for 50 epochs.

Our sequence comprises the positional coordinates of nodes, node indices, and the positional coordinates of Bézier control points. We employ a discretization resolution of 0.5m, with the node coordinates range set between 0 and 200. To distinguish the meanings of different tokens and to assign varying levels of importance, we introduce a shift to the node indices and Bézier control point coordinates. Consequently, the range for node indices extends from 200 to 350, while the range for Bézier point coordinates spans from 350 to 570.
For the special tokens introduced in Figure~\ref{example}, we omitted the representations for $<None>$ and $<From>$ to make the sequence more concise. And $<to>$,  is denoted as 571, and $<SEP>$ is represented as 572. Other special tokens, such as $<EOS>$(end token of sequence),$<BOS>$ (beginning token of sequence), and $<N/A>$ (padding token), are represented as 573, 574, and 575, respectively.

We believe that the accuracy of edges relies on the correct prediction of points, so the positions of points should carry more weight than those of edges. Therefore, we designed our model such that the tokens representing point positions, which fall within the range of 0 to 200, have a weight of 2 in the cross-entropy loss. Meanwhile, the loss weight for the remaining tokens is set to 1.

\begin{table}[t]  
 
\centering  
 \fontsize{8.5}{11}\selectfont 
 \setlength{\tabcolsep}{1.0mm}
\begin{tabular}{cc|ccc|ccc}
\toprule
\multicolumn{2}{c|}{}                                                               & \multicolumn{3}{c|}{Landmark}                                         & \multicolumn{3}{c}{Reachability}                                      \\
\multicolumn{2}{c|}{\multirow{-2}{*}{Methods}}                                      & L-P           & L-R           & \cellcolor[HTML]{EFEFEF}L-F           & R-P           & R-R           & \cellcolor[HTML]{EFEFEF}R-F           \\ \midrule
\multicolumn{1}{c|}{}                          & TopoNet~\cite{TopoNet}             & 52.5          & 47.1          & \cellcolor[HTML]{EFEFEF}49.6          & 46.9          & 10.8          & \cellcolor[HTML]{EFEFEF}17.5          \\
\multicolumn{1}{c|}{}                          & LaneGAP~\cite{LaneGAP}             & 49.9          & 57.0          & \cellcolor[HTML]{EFEFEF}53.2          & 74.1          & 34.9          & \cellcolor[HTML]{EFEFEF}47.5          \\
\multicolumn{1}{c|}{}                          & RNTR~\cite{roadnet-iccv}           & 57.1          & 42.7          & \cellcolor[HTML]{EFEFEF}48.9          & 63.7          & 45.2          & \cellcolor[HTML]{EFEFEF}52.8          \\
\multicolumn{1}{c|}{}                          & LaneGraph2Seq*~\cite{roadnet-aaai} & 46.9          & 43.7          & \cellcolor[HTML]{EFEFEF}45.2          & 63.7          & 36.3          & \cellcolor[HTML]{EFEFEF}46.2          \\

\multicolumn{1}{c|}{\multirow{-5}{*}{Default}} & SeqGrowGraph                       & \textbf{63.6} & \textbf{50.8} & \cellcolor[HTML]{EFEFEF}\textbf{56.4} & \textbf{75.5} & \textbf{61.4} & \cellcolor[HTML]{EFEFEF}\textbf{67.8} \\ \midrule
\multicolumn{1}{c|}{}                          & RNTR*    ~\cite{roadnet-iccv}                            & 39.9          & 31.0          & \cellcolor[HTML]{EFEFEF}34.9          & 63.3          & 24.9          & \cellcolor[HTML]{EFEFEF}35.7          \\
\multicolumn{1}{c|}{}                          & LaneGraph2Seq*  ~\cite{roadnet-aaai}                  & 21.9          & \textbf{39.9} & \cellcolor[HTML]{EFEFEF}28.2          & 46.3          & 11.0          & \cellcolor[HTML]{EFEFEF}17.8          \\

\multicolumn{1}{c|}{\multirow{-3}{*}{PON}}     & SeqGrowGraph                       & \textbf{43.5} & 33.3          & \cellcolor[HTML]{EFEFEF}\textbf{37.7} & \textbf{63.6} & \textbf{36.9} & \cellcolor[HTML]{EFEFEF}\textbf{46.7} \\ 
\bottomrule
\end{tabular}
\caption{
Comparison of our method with leading approaches. This evaluates  Landmark Precision-Recall and Reachability Precision-Recall on the default and PON~\cite{pon} split of nuScenes. Results marked with * are reproduced using the official code, while other results are obtained from their respective papers or released checkpoints.
}
\label{tab1}
\end{table}

\begin{table}[t] 
\centering  
 \fontsize{8.5}{11}\selectfont 
 \setlength{\tabcolsep}{1.7mm}
\begin{tabular}{c|ccc|ccc}
\toprule & \multicolumn{3}{c|}{Landmark}                              & \multicolumn{3}{c}{Reachability}                            \\
\multirow{-2}{*}{Methods} &  L-P & L-R & \cellcolor[HTML]{EFEFEF}L-F  &  R-P  &  R-R & \cellcolor[HTML]{EFEFEF}R-F  \\ \midrule
RNTR*~\cite{roadnet-iccv}          & 50.7          & 29.4          & \cellcolor[HTML]{EFEFEF}37.2          & 68.1          & 29.6          & \cellcolor[HTML]{EFEFEF}41.3          \\
LaneGraph2Seq*~\cite{roadnet-aaai} & 60.6          & 48.7          & \cellcolor[HTML]{EFEFEF}53.9          & 75.5          & 37.8          & \cellcolor[HTML]{EFEFEF}50.4          \\
SeqGrowGraph                       & \textbf{64.1} & \textbf{50.4} & \cellcolor[HTML]{EFEFEF}\textbf{56.4} & \textbf{77.7} & \textbf{43.4} & \cellcolor[HTML]{EFEFEF}\textbf{55.7} \\ \bottomrule
\end{tabular}
\caption{
Comparison of our method with leading approaches on Argoverse 2 , based on  Landmark Precision-Recall and Reachability Precision-Recall. Results marked with * are reproduced using the official code.
}
\label{tab2}
\end{table}

\begin{table}[t] 
\centering
 \fontsize{8.5}{11}\selectfont 
 \setlength{\tabcolsep}{1.3mm}
\begin{tabular}{c|cc
>{\columncolor[HTML]{EFEFEF}}c |cc
>{\columncolor[HTML]{EFEFEF}}c |
>{\columncolor[HTML]{EFEFEF}}c }
\toprule
Methods                           & M-P           & M-R           & M-F           & C-P           & C-R           & C-F           & Det.          \\ \midrule
LaneGraph2Seq~\cite{roadnet-aaai} & 64.6          & 63.7          & 64.1          & 69.4          & \textbf{58.0} & 63.2          & 64.5          \\
SeqGrowGraph                      & \textbf{71.2} & \textbf{70.2} & \textbf{70.6} & \textbf{76.2} & \textbf{58.0} & \textbf{65.8} & \textbf{67.2} \\\bottomrule
\end{tabular}
\caption{
Comparison of our method with leading approaches on nuScenes , based on metrics including mean precision (M-P), recall (M-R),  F1-score (M-F), and connectivity precision (C-P),   recall (C-R), F1-score (C-F), and detection ratio. Results  are obtained from the original paper. 
}
\label{tab3}
\end{table}


\begin{figure}[htbp]
  \centering
  \includegraphics[width=0.9\linewidth]{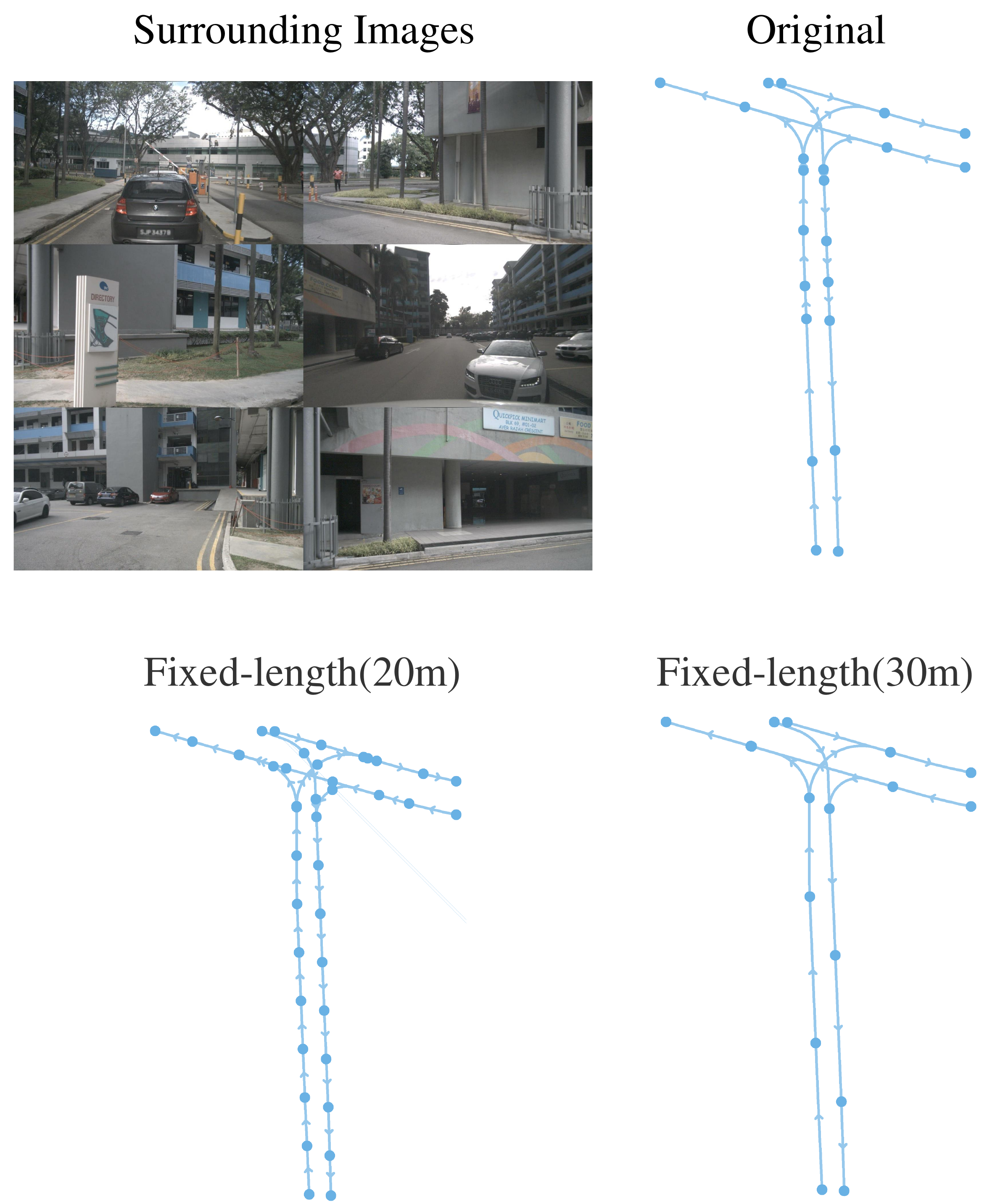}
  \caption{Various definitions of centerline split have been considered in this study. We have re-segmented the original centerline from the nuScenes dataset to enhance its reasonableness and learning accessibility.
Fixed-Length : The centerline between intersection points is partitioned at consistent intervals along the driving direction. In the provided example, these intervals are set at 20 meters and 30 meters.
By refining the split using these methods, the learning process is facilitated, thereby yielding more accurate and interpretable results of lane graphs.}
\label{gt}
\end{figure}

\subsection{Metric}

To ensure a fair comparison, we utilize the metrics proposed by RNTR~\cite{roadnet-iccv}, which encompass Landmark Precision-Recall and Reachability Precision-Recall. Landmark Precision-Recall evaluates the algorithm's accuracy in predicting the locations of key centerline points, such as segmented, fork, and merge points on lane lines. Meanwhile, Reachability Precision-Recall assesses the algorithm's capability to predict connectivity between these landmark nodes, indicating whether roads exist between them. Together, these metrics offer a comprehensive evaluation of the lane graph.

In addition, we employ metrics introduced by STSU~\cite{STSU}, including Mean Precision, Mean Recall, and Mean F1-score of centerlines, derived from precision-threshold and recall-threshold curves, alongside their F1-scores. These metrics focus on assessing the accuracy of centerline position matching, without accounting for missing predictions, thereby requiring the detection ratio to evaluate the presence of centerlines. Moreover, Connectivity Precision,  Recall, and F1-score ratios evaluate the connectivity between centerlines, primarily focusing on centerline positions and not considering crucial landmark nodes, like junction points.

\subsection{Comparison}
We compare our approach with state-of-the-art methods using the aforementioned metrics. 

Table~\ref{tab1} compares the results of our method with state-of-the-art methods on  nuScenes  using both the default and PON~\cite{pon} splits. Our method outperforms existing algorithms in terms of F1-score for both landmarks and reachability. When there is no lane overlap between the training and test sets, as is the case with the PON split, the performance of all methods significantly decreases, but SeqGrowGraph remains in a leading position.
Table~\ref{tab2} compares our method with leading methods on the larger dataset AV2 , demonstrating the superiority of our approach.
Table~\ref{tab3} presents a comparison between our method and LaneGraph2Path concerning centerline-related metrics on nuScenes. It is evident that our method still outperforms the comparison method.
This proves that modeling sequences in a manner similar to human language effectively harnesses the seq2seq model's capabilities in understanding and reasoning.

\begin{table}[t] 
 \fontsize{8.5}{11}\selectfont 
 \setlength{\tabcolsep}{2.3mm}
\begin{tabular}{c|ccc|ccc} \toprule
& \multicolumn{3}{c|}{Junction Landmark}                                & \multicolumn{3}{c}{Junction Reachability}                    \\
\multirow{-2}{*}{Segment Length} & L-P           & L-R           & \cellcolor[HTML]{EFEFEF}L-F           & R-P           & R-R           & \cellcolor[HTML]{EFEFEF}R-F  \\ \midrule
Original                       & 68.6          & 52.6          & \cellcolor[HTML]{EFEFEF}59.5          & 69.4          & 56.4          & \cellcolor[HTML]{EFEFEF}62.2 \\
20m                            & \textbf{70.5} & 52.5          & \cellcolor[HTML]{EFEFEF}60.2          & 70.6          & 57.6          & \cellcolor[HTML]{EFEFEF}63.4 \\
30m                            & 69.3          & 52.3          & \cellcolor[HTML]{EFEFEF}59.6          & 70.8          & 58.1          & \cellcolor[HTML]{EFEFEF}63.8 \\
40m                            & 69.7          & \textbf{53.3} & \cellcolor[HTML]{EFEFEF}\textbf{60.4} & \textbf{71.9} & \textbf{58.6} & \cellcolor[HTML]{EFEFEF}\textbf{64.6} \\
\bottomrule     
\end{tabular}
\caption{
Training results under different segmentation ways. We resegment the centerline split to accurately and comprehensively represent the lane graph information by re-segmenting the centerlines between junctions based on length. 
}
\label{tab4}
\end{table}

\subsection{Centerline re-segmentation}
In the original dataset, the division of map elements is based on map rules, but these division points often lack distinct visual features. For example,  even a straight road might have multiple points that split a centerline into several segments, and these division points may not have any visual features that can be easily identified.  Such data can disrupt the model's learning process and impact its ability to accurately interpret the rules. This issue is also noted in~\cite{OpenLane-V2}, and our tasks  observe instability in the position of nodes as well. To address this, when a single topological node  is referred to as a continuous node that only connects two centelines, we remove it and merge the centerlines. 
Our method fundamentally ensures that centerlines are segmented exclusively at junctions, including  forks, and merge points, which constitute the topological nodes.

When topological points are restricted to intersections, forks, and merges, the centerline may not be accurately represented by a quadratic Bézier function. To address this, we introduce additional topological points along each centerline and subsequently segment it. This segmentation is performed by dividing the centerline based on length. Figure~\ref{gt} shows the examples when the intervals are set at 20 meters and 30 meters.

We evaluate the performance of our method under specific settings by adjusting the metrics to accommodate different ground truth standards. The evaluation framework remains aligned with Landmark Precision-Recall and Reachability Precision-Recall, while incorporating data post-processing. For both the ground truth and predictions, we remove continuous nodes and merge connected centerlines. This revised metric, termed Junction Landmark Precision-Recall and Junction Reachability Precision-Recall, emphasizes the accuracy of junction locations and the connectivity between junctions. 

Table~\ref{tab4}  presents a comparison between the re-segmented centerline and the original ground truth. It can be observed that the metrics retrained after re-segmentation are all higher than those of the original split way. This suggests that we should adopt a similar standard when constructing ground truth for lane graphs, which could provide insights for research on lane graphs.

\label{Centerline re-segmentation}

\subsection{Visualization}

\begin{figure}[htbp]
 \centering
\includegraphics[width=0.9\linewidth]{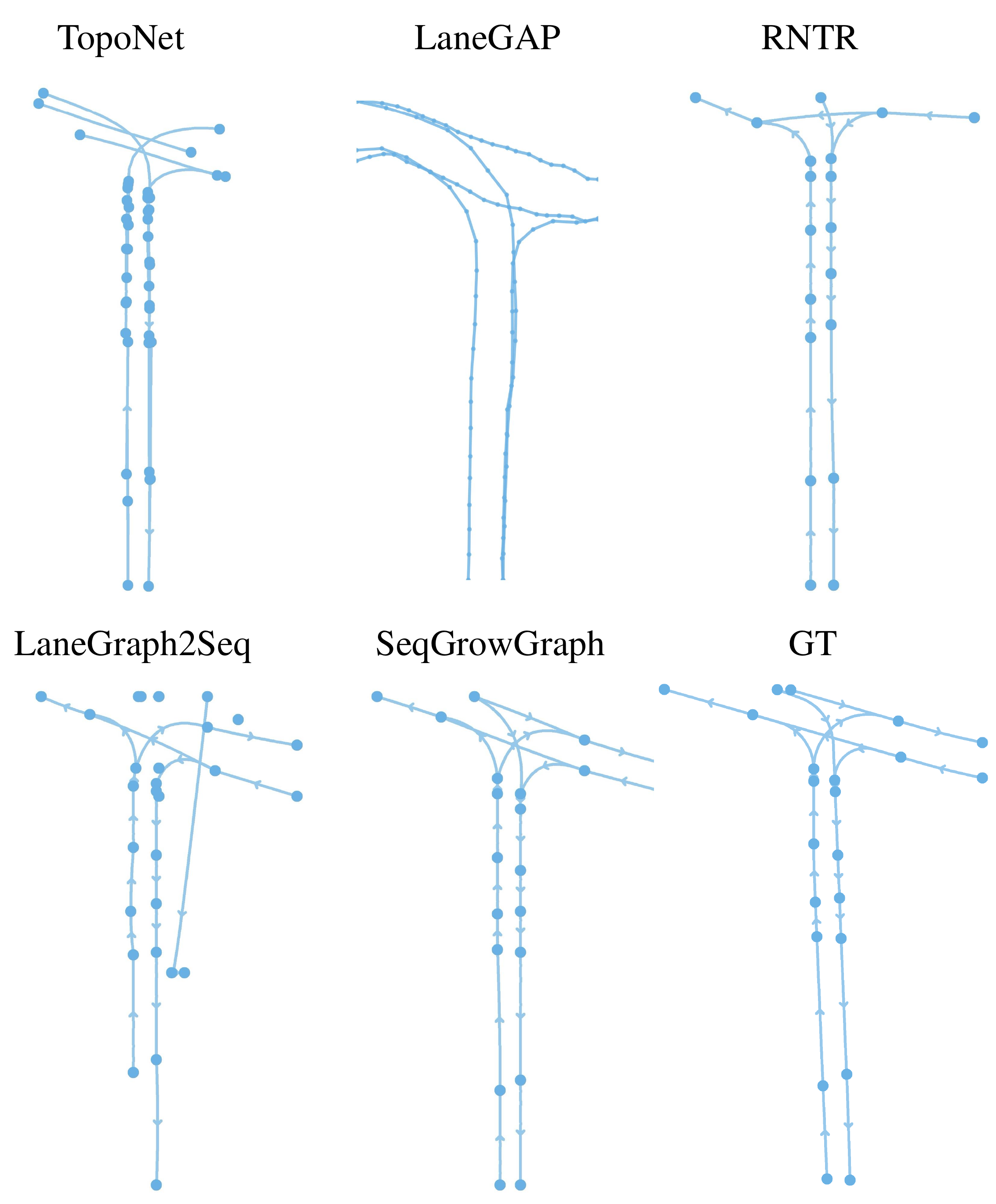}
   \caption{We visualize and compare the final results of various algorithms on the nuScenes dataset.}
\label{compare}
\end{figure}

We conduct a visual comparison of the final results produced by various algorithms on the nuScenes dataset, as illustrated in Figure~\ref{compare}. This comparison reveals several limitations inherent in previous methods: TopoNet~\cite{TopoNet} often produces centerlines that are misaligned with each other, failing to provide a continuous centerline representation. LaneGAP~\cite{LaneGAP} tends to generate duplicate centerlines and may erroneously merge adjacent centerlines, leading to inefficiencies and inaccuracies. RNTR~\cite{roadnet-iccv} demonstrates instability in its modeling capabilities, compromising the reliability of the resulting lane graphs. Furthermore, LaneGraph2Path~\cite{roadnet-aaai} may result in a decoupling between points and edges, undermining the structural integrity of the graphs. In contrast, our algorithm not only provides stable and robust modeling but also effectively merges redundant points present in the ground truth data, thereby achieving superior results in constructing lane graphs. This advancement highlights the effectiveness and reliability of our approach in overcoming these challenges.

\subsection{Ablation study}

We conduct ablation experiments on the nuScenes dataset, which includes examining the impact of loss weight on point positions and the order of subgraphs.

Higher positional weight is applied to tokens ranging from 0 to 200 because these correspond to the positions of intersection points. Accurate descriptions of these intersection positions are vital for correctly depicting the topology and centerline locations between intersections. Table~\ref{tab5} illustrates that the modeling method remains relatively stable across different weights, with a slight improvement when the weight is set to 2, highlighting the robustness of the algorithm.

The potential order in which each intersection point can be traversed to construct the corresponding subgraph is also investigated. In this context, $Center$ refers to starting from the lane graph's center and proceeding from the closest to the farthest intersection based on their distance from the center. $Coord$ denotes traversal based on the coordinates of the intersection points, specifically according to the x and y values. $BFS$ stands for breadth-first search traversal, and $DFS$ indicates depth-first search traversal. It can be observed that DFS performs the best, which suggests that depth-first search helps capture the hierarchical relationships between intersections.

\begin{table}[t] 
 \fontsize{8.5}{11}\selectfont 
 \setlength{\tabcolsep}{3mm}
\begin{tabular}{c|ccc|ccc} \toprule
& \multicolumn{3}{c|}{Landmark}                                         & \multicolumn{3}{c}{Reachability}                                      \\
\multirow{-2}{*}{Weight} & L-P           & L-R           & \cellcolor[HTML]{EFEFEF}L-F           & R-P           & R-R           & \cellcolor[HTML]{EFEFEF}R-F           \\ \midrule
1                        & 62.6          & 48.6          & \cellcolor[HTML]{EFEFEF}54.7          & 74.4          & 60.2          & \cellcolor[HTML]{EFEFEF}66.5          \\
2                        & \textbf{63.6} & \textbf{50.8} & \cellcolor[HTML]{EFEFEF}\textbf{56.4} & \textbf{75.5} & \textbf{61.4} & \cellcolor[HTML]{EFEFEF}\textbf{67.8} \\
3                        & 62.5          & 49.6          & \cellcolor[HTML]{EFEFEF}55.3          & 75.4          & 61.3          & \cellcolor[HTML]{EFEFEF}67.6          \\
4                        & 63.0          & 50.2          & \cellcolor[HTML]{EFEFEF}55.9          & 75.3          & 61.2          & \cellcolor[HTML]{EFEFEF}67.5   \\
\bottomrule     
\end{tabular}
\caption{
Ablation study on different loss weights assigned to token of point coordinates conducted using the nuScenes dataset.  The weights for the remaining tokens are set to 1. 
}
\label{tab5}
\end{table}
\vspace{-0.3cm}

\begin{table}[]
 \fontsize{8.5}{11}\selectfont 
 \setlength{\tabcolsep}{3mm}
\begin{tabular}{c|ccc|ccc} \toprule
 & \multicolumn{3}{c|}{Landmark}                                         & \multicolumn{3}{c}{Reachability}                                      \\
\multirow{-2}{*}{Order} & L-P           & L-R           & \cellcolor[HTML]{EFEFEF}L-F           & R-P           & R-R           & \cellcolor[HTML]{EFEFEF}R-F           \\ \hline
$Center$                & 59.0          & 47.1          & \cellcolor[HTML]{EFEFEF}52.4          & 73.3          & 56.3          & \cellcolor[HTML]{EFEFEF}63.7          \\
$Coord$                 & 60.1          & 48.6          & \cellcolor[HTML]{EFEFEF}53.7          & 73.7          & 57.2          & \cellcolor[HTML]{EFEFEF}64.4          \\
$BFS$                   & 62.6          & 49.2          & \cellcolor[HTML]{EFEFEF}55.1          & 75.3          & 61.4          & \cellcolor[HTML]{EFEFEF}67.7          \\
$DFS$                   & \textbf{63.6} & \textbf{50.8} & \cellcolor[HTML]{EFEFEF}\textbf{56.4} & \textbf{75.5} & \textbf{61.4} & \cellcolor[HTML]{EFEFEF}\textbf{67.8} \\\bottomrule  
\end{tabular}
\caption{
Ablation study on the ordering of subgraph descriptions conducted using the nuScenes dataset. $Center$ pertains to initiating from the central point of the lane graph and navigating towards intersections in order of their increasing distance from this center. $Coord$ involves traversing based on the coordinates of intersection points, specifically ordered by their x and y values. $BFS$  and $DFS$  refers to breadth-first  traversal and depth-first   traversal.
}
\label{tab6}
\end{table}

\label{sec:experiments}

\section{Conclusion}

In this paper, we introduce SeqGrowGraph, an innovative framework for autoregressive modeling of lane graphs that converts lane graphs into sequences via incremental local growth, facilitating a comprehensive global understanding. Our method enables flexible representation of complex road structures, including loops and bidirectional lanes, which pose challenges for traditional DAG-based approaches. Extensive experiments conducted on the nuScenes and Argoverse 2 datasets demonstrate superior performance in topological accuracy and graph completeness compared to existing detection-based and generation-based methods. SeqGrowGraph advances lane graph construction by bringing human-like structural reasoning and sequence modeling, thereby providing accurate centerlines for autonomous driving in dynamic environments.
\label{sec:conclusion}

{\small
\bibliographystyle{ieeenat_fullname}
\bibliography{11_references}
}


\end{document}